# Performing an inductive Thematic Analysis of semi-structured interviews with a Large Language Model

## An exploration and provocation on the limits of the approach

**Stefano De Paoli** – Abertay University – s.depaoli@abertay.ac.uk

## Abstract

Large Language Models (LLMs) have emerged as powerful generative Artificial Intelligence solutions. This paper presents results and reflections of an experiment done with the LLM GPT 3.5-Turbo to perform an inductive Thematic Analysis (TA). Previous research has worked on conducting deductive analysis. Thematic Analysis is a qualitative method for analysis commonly used in social sciences and it is based on interpretations by the human analyst(s) and the identification of explicit and latent meanings in qualitative data. The paper presents the motivations for attempting this analysis, it reflects on how the six phases to a TA proposed by Braun and Clarke can partially be reproduced with the LLM and it reflects on what are the model's outputs. The paper uses two datasets of open access semi-structured interviews, previously analysed by other researchers. The first dataset contains interviews with videogame players, the second is a dataset of interviews with lecturers teaching data science in a University. This paper used the analyses previously conducted on these datasets to compare with the results produced by the LLM. The results show that the model can infer most of the main themes from previous research. This shows that using LLMs to perform an inductive TA is viable and offers a good degree of validity. The discussion offers some recommendations for working with LLMs in qualitative analysis.

## 1. Introduction

The problem investigated in this paper is whether we can use the Large Language Model (LLM) GPT (3.5-Turbo), to perform an inductive Thematic Analysis (TA) of semi-structured interviews. This investigation is exploratory and does not seek to establish formal procedures for doing a TA with LLMs. I am interested in learning something either about the about the method or about how LLMs can be used in qualitative analysis. LLMs are generative Artificial Intelligences (AI), working with natural language processing (NLP), trained on massive amounts of textual data. They use advanced machine-learning neural networks with the capacity to learn from data and improvise responses. LLMs can produce meaningful conversation and interactions with humans when prompted and solve tasks. LLMs do not produce human level thinking, for example they do not possess an idea of the world like humans do (Hao et al., 2023). For Floridi (2023, p. 14) LLMs: "can do statistically—that is, working on the formal structure, and not on the meaning of the texts they process—what we do semantically". LLMs operate on the language not with an understanding of its meaning like humans, but from a structural perspective, where words are seen in their numerical representation and sentences are built using structural and probabilistic components of language.

This leads to the provocative nature of my inquiry, which relates to the analysis social scientists operate within qualitative research, considered inductive, interpretative as opposed to the positivist approach focused on deduction or logic. The father of Sociology Max Weber, for example noted that the goal of social scientists is to interpret the meaning of social action for providing a casual explanation. This interpretation is subjective, and the analysts use their own sensibilities and knowledge. For example, TA focuses on the identification of "patterns of meaning": themes in data. Braun & Clarke (2006, p. 81 – emphasis added) argued that TA: "can be an essentialist or realist method, which reports **experiences, meanings and the reality of participants,** or it can be a constructionist method, which examines the ways in which **events, realities, meanings, experiences** and so on are the effects of a range of discourses operating within society". Qualitative analysis thus works on meanings and interpretation, whereas LLMs work on structural and probabilistic elements of language.

TA is a flexible approach to qualitative analysis and lends itself to experimentation with LLMs. Braun & Clarke (2006) famously stipulated that researchers should operate six, inter-related phases: "(1) familiarising yourself with your data; (2) generating initial codes; (3) searching for themes; (4) reviewing themes; (5) defining and naming themes; (6) producing the report". Because of this step-by-step process, I believe it is possible to work separately on each phase and see if an LLM can perform an inductive TA. Inductive TA is created without having pre-determined codes and themes: the analysts ground the codes strictly to the data and do not fit the data into any pre-existing frame (Nowell et al., 2017). The inductiveis different and separate from a deductive approach, where analysts attribute data extracts to a pre-defined grid of analysis. Said otherwise, the two terms describe very different relations between theory/concepts and data: in the deductive the analysts rely on a theoretical framework or defined categories to interpret the data, whereas in the inductive approach the analysts ignore the theory and openly explore themes or concepts (see for a discussion Kennedy & Thornberg, 2018). Developing an inductive approach with an LLM is an interesting problem, because it puts the LLM in the position to creatively derive codes/themes directly from the data, without any pre-defined analytical frame.

## 2. Literature Review

There has been significant hype around LLMs such as ChatGPT including in research and academia. Data analysis is a potential area of application for LLMs, and literature often relates to the field of data science (Hassani and Silva, 2023). For example, ChatGPT has been used for exploratory data analysis (Sanmarchi et al. 2023), visualization or preprocessing (Kurniadi et al., 2023). The impact on research (van Dis et al., 2023) and research priorities are also discussed in publications, including in healthcare (Sallam, 2023), financial research (Dowling and Lucey, 2023), or ethics of publishing (Lund et al., 2023b) to mention few examples. Others focused on the application of ChatGPT such as in marketing research (Peres et al., 2023), or education (Rahman, & Watanobe, 2023). There already are literature reviews for example in education (Lo, 2023), healthcare (Sallam, 2023), or business (George, 2023). Readers can consult these for an overview, but the literature field is in constant evolution.

Some similarities could be drawn between the use of LLMs and previous work using "traditional" supervised/unsupervised machine-learning for qualitative analysis such as qualitative coding (e.g. Renz, 2018), grounded theory (Nelson, 2020), content analysis (see e.g. Renz et al., 2018), or TA (Gauthier & Wallace, 2022) ) at least insofar as activities related to analysis are delegated to algorithms. In a review of the gaps in Computational Text Analysis Methods especially in the social sciences, Baden et al. (2016) noticed that there has been primacy of technological solutions over reflecting on the validity of using computational resources for analysis, and that at the same time the application of computational resources to text analysis tends to be narrow, focusing on a single aspect such as the sentiment. There has been a limited uptake of machine-learning in qualitative analysis, also because of social scientists' lack of computing skills.

We should consider that using LLMs for qualitative analysis is a subject in its *statu nascendi,* in the state of being born. Therefore, there is very limited literature available. To the author's knowledge there are only two scientific papers on the use of LLMs for qualitative data analysis (Xiao et al., 2023; Gao et al. 2023). An additional online post is also worth discussing (Schiavone et al., 2023), due to its findings. The first two contributions also place emphasis on the labor-intensive process of qualitative coding of large datasets. Both papers promote the use of LLMs to automate the qualitative coding process. Neither engages substantially with the problem of 'interpretation' which is part of qualitative analysis. Xiao et al. (2023) focus on two aspects: the agreement of LLMs deductive coding with coding done by human analysts, and how the design of the prompt (i.e. what is asked to the LLM) impacts the analysis. They deploy LLMs for deductive coding, which in their paper means assigning predefined labels to text. They found that "it is feasible to use GPT-3 with an expert-developed codebook for deductive coding." (p. 76) and that a "codebook-centered design performs better than the example centered designs" (p. 77), where the prompt is based on the actual codes, rather than on examples of using the codebook. They use Cohen's Kappa to measure the inter-reliability between the analyst and the model, however this measure can only be used when different analysts use the same codes on the same material. Gao et al. (2023) focus on the creation of a support tool for collaborative coding. They connect the use of LLM to social sciences and follow the approach proposed by Richards and Hempill (2018), focusing on "open coding", "iterative discussion", and "codebook development". The research is also supported by user evaluation, which contributes to establishing some agreement across the scientific community. Gao et al. (2023) suggest relevant implications, including the use of LLMs as helpers more than as replacement of analysts, the use of the results for discussion amongst the research team and as basis for refinement. These suggestions fall in the Human-AI collaboration in qualitative analysis suggested by Jiang et al. (2021). Lastly, although not a scientific paper, the online post by Schiavone et al. (2023) deserves mention as it reports the results of a TA conducted on a small set of online user comments. The authors operated a TA both manually and then with ChatGPT to assess the reciprocal inter-reliability of human-to-human and human-to-LLM showing that the Cohen's Kappa metric appears largely similar in both cases (around circa 0.7).

In summary, Xiao et al. (2023) and Gao et al. (2023) offer interesting insights, e.g. the role of prompts in relation to the outputs, or the use of models as support of analysist. I propose, however, a different approach. Firstly, to work on an inductive TA process and understand if something satisfactory can be extracted from the model. Secondly, I come from a perspective of social sciences and my goal is not building tools, but to reflect on the methodological aspects of doing TA with LLMs. Thirdly, I propose working on qualitative interviews, whereas they worked with secondary data.

## 3. Design of the experiment and LLM use

Using existing open access semi-structured interviews previously analysed by other researchers, I will attempt at re-analysing the data with the LLM (GPT3.5-Turbo) to see what output it generates, in terms of codes and themes. I will then compare these themes with the original analysis to reflect on whether a LLM TA has some validity. The operationalisation of codes and themes I use follows exactly Braun and Clarke (2006) own definitions (see later). For the comparison, I will assess if the model can generate similar names for themes and if the themes descriptions are similar or match those of the original researchers. I will only consider phases 1-5 of a TA. Phase 6 relates to writing up the results, and as there is discussion on using LLMs for writing academic publications, this will not be attempted here.

I will use two datasets of semi-structured interviews, selected because of the following: 1) they are open access with creative commons licenses; 2) because of this, they are anonymised and do not raise specific ethical concerns; 3) we have documents reporting the analysis and can draw a comparison; 4) they are contained in size, and this is beneficial since GPT has inherent limits related with text processing.

The first dataset is the player interviews (n=13, young people between 18-26 years of age) from the project gaminghorizon, an EU funded "project that explored the role of video games in culture, economy and education"[1]. Players were one of the stakeholders' groups of the project alongside e.g. educators or policymakers. The dataset is available from zenodo (Perrotta et al., 2018). We will call this the 'gaming' dataset. There is an associated report with the results of a TA (Persico et al., 2017a), and a literature review (Persico et al., 2017b) where the analysis framework is defined. Albeit the analysis was done with a deductive approach, using the results from the reports it will be possible to identify whether there are similarities or differences with what GPT can produce inductively. The second dataset (Curty et al. 2022) is related with the project "Teaching undergraduates with quantitative data in the social sciences at University of California Santa Barbara", it comprises 10 interviews with instructors/lecturers "who use quantitative data to teach undergraduate courses" at the University. We will call this the 'teaching' dataset. There s an associated report (Curty et al. 2021), with the results of inductive coding.

### 3.1 API, Prompt and Response

The experimentation was conducted with the OpenAI API, which allows to connect to GPT3.5-turbo (https://platform.openai.com/docs/introduction/key-concepts) via python scripts (a script is a small computer program for a computer to perform). In the following, the python scripts will not be discussed in detail, as they are doing basic operations on the data. However, we will discuss in detail the prompts, with the instructions given to the LLM. We must note also that the model is a black-boked AI and we do not know what operation it does when requested to perform a prompt, due to the complexity of the underlying algorithms (see e.g. Rai, 2020 for a discussion on black-boxed AI), and the proprietary nature of the software. What we know, as users, is that we give the model a textual input (prompt), and we will receive a textual output (response). A prompt is the set of textual instructions given to the model, whereas the response is the model's output based on its 'interpretation' of the prompt. Prompting is when "a pre-trained language model is given a prompt (e.g. a natural language instruction) of a task and completes the response without any further training or gradient updates to its parameters." (Wei et al., 2022, p. 3). For example, in ChatGPT (https://chat.openai.com/), one example is as in Figure 1. The user types the prompt "write me the names of the 3 most important Italian poets", and the model responds. We do not know what the model does with our prompt, but we see its output and can assess it. This is exemplified in Figure 2.

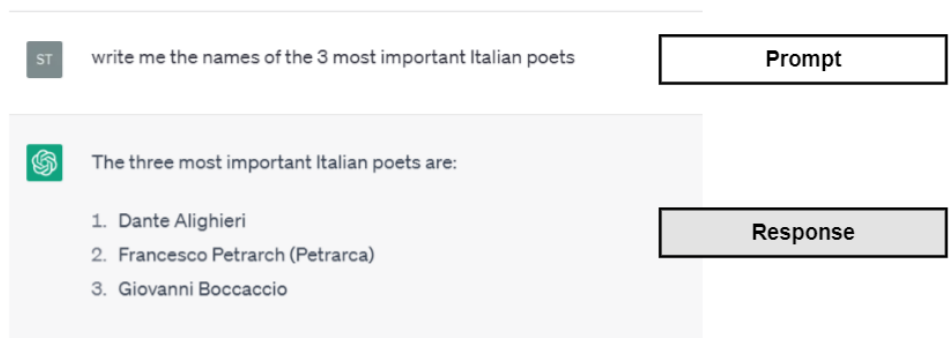

**Figure 1** – Prompt-response example from ChatGPT

---

[1] https://www.gaminghorizons.eu/

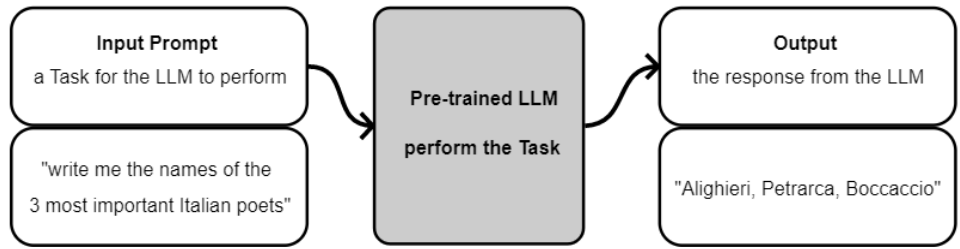

**Figure 2** – LLM prompting and response, simplified workflow

The API works in a similar way, but it is possible to use this in conjunction with a scripting language and this allows additional data manipulations and parsing. Using the API, the script-code of the previous example will look like the one in Figure 3.

```
In [1]: import openai

In [2]: k=KEY

        openai.api_key = k

In [3]: def get_completion(prompt, model="gpt-3.5-turbo"):
            messages = [{"role": "user", "content": prompt}]
            response = openai.ChatCompletion.create(
                model=model,
                messages=messages,
                temperature=0, # this is the degree of randomness of the model's output
            )
            return response.choices[0].message["content"]
```
Function for using the LLM with the prompt

```
In [4]: prompt = f"""
        write me the names of the 3 most important Italian poets`
        """
        response = get_completion(prompt)
        print(response)

        1. Dante Alighieri
        2. Francesco Petrarca
        3. Giovanni Boccaccio
```
Prompt

Response

**Figure 3** – Python script, prompt and response replicating the example of Figure 1

The script requires importing the OpenAI library ([1]) which allows one to connect to the LLM API. The connection requires a secret key (**KEY**) to access the model ([2]). The function **get_completion** ([3]) calls the model (gpt-3.5-turbo) to work on the prompt (in [4]) and returns a response (last two lines in [4]). This function and the prompt are sufficient for using the API, and the response from the LLM in figure 3 is the same as in figure 1. The building of prompts is a critical moment of using the model, as the user keeps refining them to produce the desired output (Zhou et al., 2022). Indeed, even slight variations in the prompt wording may yield different results, due to the probabilistic nature of LLMs responses. The concept "prompt engineering" (see e.g. Wei et al., 2022), encapsulates the testing needed to provide a clear set of instructions to the model. Also note the parameter **temperature** (T) within the function get_completion (in [3]), which relates with the 'randomness' and 'creativity' of the model. The temperature accepts values between 0 and 2. This implies that by running the same prompt with temperature at 0 (no-randomness-deterministic) the model should reproduce the same output. Instead, higher values will increase response variability. For example, the above prompt with T at 1, gave: Dante Alighieri, Francesco Petrarca and Ludovico Ariosto. Running this again might lead to further different responses.

Lastly, there is an important difference between the webchat and the API, in the second case it is possible to have the LLM do operations responding to the prompt on textual material, such as interviews or other natural documents. Figure 4 exemplifies this point: with the API it is possible to pass the data we want to analyse to the LLM, and have the model perform a task.

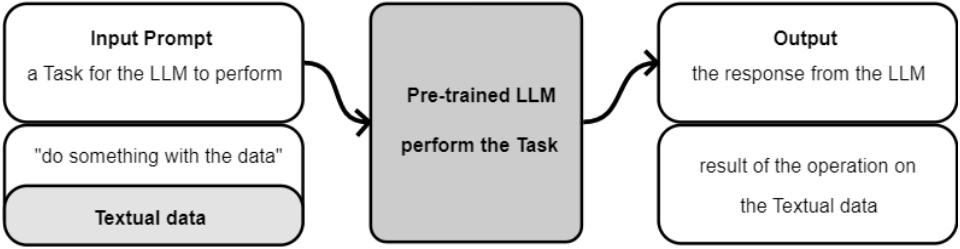

**Figure 4** - Workflow with data in the prompt via API

### 3.2 Tokens and memory

LLMs have a limit related to how many tokens they can process at any one time. A token roughly equates to one word. At the time of writing the limit for GPT3.5-Turbo is 4097 tokens including both the prompt and response. This limit impacts text processing. For instance, the interviews from the 'gaming' dataset are between 5000 to 9000 words. Therefore, interviews must be divided into smaller chunks to be processed one at a time. We have also to consider that interviews will need to be part of the prompt, and therefore chunks must be relatively small to allow the LLM to have enough tokens to produce the response.

I wrote a script to divide the datasets into chunks of roughly 2500 tokens. This number has been selected after experimenting with higher values (e.g. 3000) which would occasionally reach the max tokens limit. With a chunking at around 2500 tokens the 'gaming' dataset resulted in 56 chunks and the 'teaching' dataset in 35 chunks. The chunks have been stored in a csv file with the structure of Figure 5.

| FileName | Interview_chunk | Tokens |
|---|---|---|
| part_0_Play_1.txt |  | 2479 |
| part_1_Play_1.txt | things but | 2513 |
| part_2_Play_1.txt | or | 2486 |
| part_3_Play_1.txt | of art is in | 1740 |

**Figure 5** – Structure of the CSV file containing the chunks

FileName is the name of the chunk, e.g. part_0_Play_1 is the first chunk of the Play_1 interview. Then there is the text of the chunk (Interview_chunk) and a column with the tokens number.

Another important aspect of the LLM is that it does not have memory, i.e. it does not remember the content of past prompts, and if these are relevant for later operations, then these will need to be passed to the model again as prompts. This is a limit and the LLM can work only one chunk at a time. Therefore, for this experimentation I assumed that the model would not 'remember' previous prompts.

## 4. Results and Observations

In this section I present the analysis done on the datasets. The methodological components of the research are presented alongside the analysis results, since the objective is to establish whether we can perform something looking like a TA with an LLM (i.e. the methods are also the results). This follows the workflow described in Figure 6. In each phase different prompts are used, and these will be detailed in the following pages. The TA is done inductively: no preexisting coding framework orexamples are given to the model, and all the codes and the subsequent themes, are entirely grounded in the LLM data 'interpretation'.

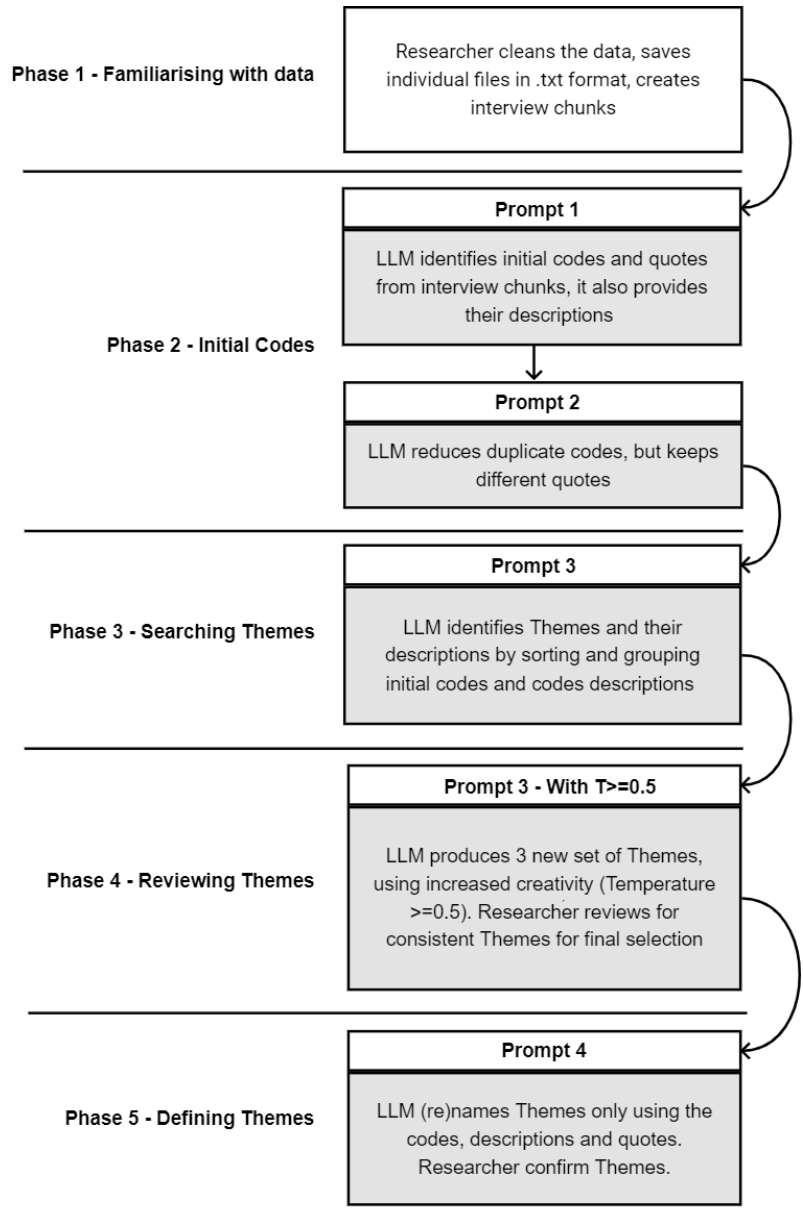

**Figure 6** – Simplified workflow for a TA with an LLM

*Phase 1: familiarising with the data*

This phase requires the researcher(s) to familiarise with the data, by e.g. transcribing interviews or reading transcripts. This is done to begin formulating insights into what the data is about. I believe that this familiarisation now cannot be performed with e.g. GPT, due to the tokens and memory limits. It may be possible with more powerful models to have the model read all the material and 'familiarise' with its contents. Nonetheless in this phase it is important for the researcher(s) to prepare the data for processing.

Due to the limit of tokens, it is useful to clean the data before the analysis. For example, upon inspection of the 'gaming' interviews, the first opening page of the transcripts was about e.g. setting up the recording. For example, part of the opening of most interviews was as follows:

> First of all, you do know you are audio recorded?
>
> Yeah.
>
> Great. Can you see in the screen I'm sharing?
>
> Yeah.
>
> Great. I will explain to you how this interview works and what we are interested in....

This is not text which has relevance for the analysis and was trimmed to reduce the number of tokens. This clean-up was done manually, removing irrelevant text at the beginning or at the end (e.g. final salutation). After this, the chunking operation previously described (see Figure 5) can be done. Additionally, it may be useful to change the format of the data. For example, the 'gaming' interviews were .rtf files and the 'teaching' were pdfs. Everything was saved as text files (.txt) for processing.

*Phase 2: generating initial codes*

The second phase of a TA is the generation of codes. For Braun & Clarke (2006), codes "identify a feature of the data (semantic content or latent) that appears interesting to the analyst" (p. 88). The analysts identify features of interest in the data that may have some meaning and about the meaning the respondents attribute to e.g. things, social relations, events under investigation. I asked the model to inductively infer 'codes' from the data, without any coding scheme using the prompt in Figure 7. In the prompt, the word 'themes' is used instead of 'codes' as it works better within the prompt. However, for clarity here we are identifying the initial 'codes'.

```python
for i in range(l):
    text = df.loc[i]['Interview_Chunk']

    prompt = f"""

    Identify up to 3 most relevant themes in the text, provide a name for each theme in no more than 3 words,
    4 lines meaningful and dense description of the theme and a quote from the respondent for each theme no longer than 7 lines

    Format the response as a json file keeping names, descriptions and quotes togeter in the json, and keep them
    together in 'Themes'.

    ```{text}```
    """

    response = get_completion(prompt)
    print(response)
```

**Figure 7**– Prompt to infer initial codes (**Prompt 1**)

The csv file with the interview chunks had been parsed earlier in a dataframe - df (a powerful data structure in python). The prompt is embedded in a "for cycle" running for the entire length (range(l)) of the dataframe (i.e. for the number of chunks). The script reads each chunk one at a time from the dataframe (df.iloc[i]['Interview_chunk']) and puts it in the variable 'text' ({text}) which is inside the prompt (i.e. Figure 4 workflow).

The prompt asks the model (using the get_completion function,) to:

1. Identify 3 of the most relevant codes for each chunk ({text})
2. Provide a 3 words name to each code
3. Provide a 4 lines description of the code
4. Store one meaningful quote for each code
5. Format each set of 3 codes, descriptions and quotes as a json fileThe number of codes inferred from each chunk is limited to 3, because in subsequent phases some of the output of this operation will need to be used in prompts, and this impacts the maximum number of tokens. When I asked the generation of 4 codes, the system did reach the tokens limit in Phase 3.

This prompt (Figure 7) produced 161 codes from the 'gaming' dataset and 101 from the 'teaching' dataset. These were stored in a csv file with the structure of Figure 8. Each code has a name, a description and a quote.

| Index | Codes | Description | Quote |
|---|---|---|---|
| 0 | Importance of Data An | The respondent considers c | I consider it one of the most important classes I teach because regardless of what career they go |
| 1 | Providing Data Sets | The respondent provides th | I have all of the data that they get. So there are specific assignments for different types of modu |
| 2 | Teaching Critical Thinki | The respondent spends a w | And I asked the students, you know, "What does this mean? Do you know what this means? Y |
| 0 | Misleading Graphs | The theme of misleading gr | And so we talked about where people place their axes and all of that... we have to assess what t |

**Figure 8** – Structure of the csv file with the codes inferred (excerpt)

Before considering the next phase, it is important to note that since the model goes through each chunk separately, it is likely that there will be repeated codes. This repetition would happen rarely when the analysis is done by a human, since if a portion of interview reminds the analyst about an already created code, then this is coded accordingly. Therefore, it is essential to reduce the codebook, by merging very similar codes, descriptions and quotes. For the 'gaming' dataset I also reduced the length of descriptions, because of the tokens limit. For the reduction of codes, I used the prompt in Figure 9.

```
prompt = f"""

Determine which items in the following list of topics are entirely unique \

Format the response as a json, grouped in 'items' and separated in keys 'topic' and 'indices'

List of topics: {", ".join(topic_list)}

"""
```

**Figure 9** – Prompt for reducing duplicate codes **(Prompt 2)**

It took several iterations to engineer this prompt, and I worked initially on making the model trying to identify similarity (e.g. "Determine if there are items which are very similar"). The response, however, was inconsistent. I would often get only the list of similar codes, but not unique codes. The output in the json file was also sometimes not well formatted. This is an example of how much prompt construction impacts the output quality. Note the prompt uses the terms items and topics, for clarity with the model. Using 'items' and 'topics' allowed me to distinguish between the content and the position of each topic in the list of topics/codes. The list of topics includes the code name, its description and the index (i.e. the dataframe row number). This prompt reduced the 'gaming' codebook from 161 to 89, and from 101 to 63 for the 'teaching' codebook.

During this reduction, I found that the model would sometimes get hallucinated and generate new code names out of the existing ones. Hallucination is a concept in machine-learning where the system produces a response which is not justified by the input. The prompt in itself also was not enough to guarantee consistency in the reduction of the duplicated codes. In experimenting with the prompt, I found that passing together with the code name, also the dataframe index would remove the hallucination. This was done by merging strings, where each topic in the list is as follows: 'code name': 'index' 'description'.

*Phase 3: searching for themes*

In this phase the focus moves toward the identification of themes - patterns encapsulating significant aspects of the data - where codes are grouped and sorted. Whilst this phase should remain generally open, I asked the model to generate a roughly equivalent number of themes to those produced by the original researchers, in order to perform a comparison later.

The original analysis of the 'gaming' dataset is presented in two project reports (Persico et al. 2017a and 2017b) and the specific themes will be presented later when an interpretation of the results will be provided. For now, it suffices to say that the researchers had 10 themes, and one of these themes also had 3 sub-themes. Therefore, using the 89 codes from Phase 2 I asked the model to generate 11 themes (a number between 10 and 13), using the prompt in Figure 10, where the list of topics is a list containing the name of the codes and the associated description (i.e. each topic has this format 'code': 'description'). Quotes were not included because of the tokens limit.

```
prompt = f"""

Determine how all the topics in the following list of topics can be grouped together, \
and topics can also be in more than one group. \

Group all the topics numbers only and provide a name and a description for each group\
Create 11 significant groups\

Display the full list

List of topics: {", ".join(topic_list)}

"""
```

**Figure 10** – The prompt used to infer themes inductively **(Prompt 3)**

The list of themes (called groups due to the prompt) with their description as provided by the model (with no changes or alterations made by the researcher) is presented in Table 1.

| Nr. | Theme | Description |
|---|---|---|
| 1 | Gaming and Education | This group includes topics related to the use of video games for education and learning, including the benefits and challenges of using games for teaching, the importance of creating education-based games, and the potential of gamification in education. |
| 2 | Ethics in Gaming | This group includes topics related to ethical issues in gaming, such as violence, diversity, monetization, piracy, and inclusion. It also includes discussions on the responsibility of developers and publishers to balance creative expression with ethical considerations. |
| 3 | Monetization in Gaming | This group includes topics related to how games make money, including free-to-play with in-game purchases, paid games, and pay-to-win models. It also includes |

| | | discussions on the ethical implications of micro-transactions and gambling-like systems. |
|---|---|---|
| 4 | Video Games as Art | This group includes topics related to the growing acceptance of video games as an artistic medium, citing examples of games that are considered high art. It also includes discussions on the combination of traditional art forms and gameplay. |
| 5 | Game Development | This group includes topics related to game development, including advice for developers, the importance of good management and communication, and the role of innovation in creating immersive experiences. |
| 6 | Representation in Gaming | This group includes topics related to representation in gaming, including diversity in race and gender, the importance of relatable characters, and the progress that has been made in the industry. |
| 7 | Mobile Gaming | This group includes topics related to mobile gaming, including its accessibility and oversaturation in the market. |
| 8 | Esports | This group includes topics related to esports, including its benefits, the importance of physical and mental health, and the need for moderation and balance. |
| 9 | Gamification | This group includes topics related to gamification, including its use for education and productivity, the need for entertainment to motivate people, and the potential for complicity in neoliberal capitalism. |
| 10 | Physical Health and Gaming | This group includes topics related to the impact of gaming on physical health, including the potential for reduced physical activity and the benefits and drawbacks of using games for physical rehabilitation. |
| 11 | Gaming Communities | This group includes topics related to gaming communities, including their inclusivity and toxicity levels, the significance of social interactions in gameplay, and tips for positive game interaction. |

**Table 1** – Themes inferred by the model, 'gaming' dataset

The analysis of the 'teaching' dataset had been done with an inductive approach, which is presented in the project report (Curty et al., 2021). The report had 5 high-level themes, and 3 sub-themes for one theme. Therefore, using the 63 codes I asked the model to generate 7 themes (a number between 5 and 8) with the same prompt. The resulting themes are presented in Table 2.

| Nr. | Theme | Description |
|---|---|---|
| 1 | Teaching Data Analysis and Interpretation | This group includes topics related to teaching students how to analyze and interpret data, including identifying good and bad graphs, understanding statistical knowledge, and teaching critical thinking about data. |
| 2 | Mentoring and Diversifying the Field | This group includes topics related to mentoring young students and making a difference in diversifying the field of data analysis. |
| 3 | Teaching GIS and Geospatial Data | This group includes topics related to teaching GIS software and geospatial data, including challenges in teaching and the practical use of the software. |
| 4 | Collaborative Learning and Interpersonal Interaction | This group includes topics related to the benefits of collaborative learning and interpersonal interaction in acquiring quantitative skills. |
| 5 | Teaching Research Methods | This group focuses on teaching research methods and the challenges of accessing and using data. The group emphasizes the importance of technical and statistical skills in survey research exercises. |

| 6 | Teaching Programming and Technical Skills | Group related to teaching students programming and technical skills, including the lack of programming classes available to non-computer science majors and the need for a quantitative social science minor or data sciences program. |
|---|---|---|
| 7 | External Support and Resources for Teaching with Data | Group related to the lack of external support and resources for teaching with data, including the need for training opportunities and a centralized resource for instructors. |

**Table 2** – Themes inferred by the model. 'teaching' dataset

*Phase 4: reviewing themes*

For Braun & Clarke (2006) this phase requires revising themes from Phase 3 and re-organise the analysis. For example, some themes may fit better as sub-themes, others are not consistent or homogeneous. This phase is probably much more strongly reliant on human interpretation than the previous two. Nonetheless, I believe it is possible to attempt to confirm which themes seem valid, rather than e.g. sub-themes or just codes or see if themes were overlooked. To approach this, I first re-built the full codebook composed of themes and underlying codes, the description of each code and all the associated quotes. An example from the 'gaming' analysis is presented in Table 3 (with just the theme and the codes).

| **Theme Example** | | **Gaming and Education** |
|---|---|---|
| Codes | 1 | Problem Solving |
| | 2 | Educational Potential |
| | 3 | Success Metrics |
| | 4 | Teaching through games |
| | 5 | Gender and Diversity in eSports[2] |
| | 6 | Practical Obligation of Games |
| | 7 | Off-the-shelf games |
| | 8 | Games in Education |

**Table 3** – Example of full theme and underlying codes

I propose that one way for operating this phase is to work with the temperature parameter, increasing the creativity of the LLM. Increasing the temperature in the python function **get_completion** (Figure 3) and running again the Prompt 3 (Figure 10) with the codes (from Phase 2), we can see if there are significant differences in the themes produced. Table 4 presents themes with a temperature of 1 for the 'gaming' dataset, on three tests. The goal is to identify consistency across themes between the ones from Phase 3 (Table 1) and the ones from Phase 4 and if there are overlooked themes or which appear less relevant. The choice of the final themes would need to rely on the sensibility of the human researcher. For the purposes of this paper, I have not made a specific choice in this phase but will use the themes from both Phase 4 and Phase 3 for a comparison with the original analysis.

| | **Theme names (in the order provided by the model, T=1)** | | |
|---|---|---|---|
| Nr | Test_1 | Test_2 | Test_3 |
| 1 | The Benefits of Gaming and Education | The Positive Impacts of Gaming | Using video games for learning and education |

---

[2] Although this code is about eSports, I believe the model included this code here as the description also reflects on career paths and career choices. Alternatively, this may be an hallucination, and potentially something to review manually.

| 2 | Ethical Issues in Gaming | Ethical Issues in Gaming | Ethical concerns in gaming |
|---|---|---|---|
| 3 | Monetization and Business Models in Gaming | Gaming and Art | Video games as art |
| 4 | Video Games as Art | Free-to-Play and Monetization models | Representations in gaming |
| 5 | Gaming and Age Restrictions | Gaming Industry and Development | Monetization and business models in gaming |
| 6 | Game Development and Management | Diversity and Representation in Gaming | The impact of gaming on physical and mental health |
| 7 | Gaming for Relaxation and Nurturing Experiences | Gaming in Education | Video games for recreation and relaxation |
| 8 | Immersive World-Building and Story-Driven Games | Identity in Gaming | Video games in sports and eSports |
| 9 | Diversity and Representation in Gaming | Gaming and Physical Health | Innovation and creativity in game development |
| 10 | eSports and Competitive Gaming | Esports | Using gamification for non-entertainment purposes |
| 11 | Gaming and Physical Health | Positive Gaming Interaction | Positive aspects of gaming |

**Table 4** – Themes generated, 'gaming' dataset T=1

We can see from the three tests some consistency between Phase 3 and Phase 4 around several key themes, which include Education, Art, Ethics, Monetisation, Esports, Physical Health, Representation among others. Based on this limited testing we can preliminarily conclude that these just mentioned may be potentially valid themes which represent the data. However, there also are a few other themes which might have been overlooked by the model in Phase 3 and which appear in Phase 4, such as Mental-Health and Age Restrictions.

What is presented in Table 5, is the same operation on the 'teaching' dataset.

| | Theme names (T=1) | | |
|---|---|---|---|
| Nr. | Test_1 | Test_2 | Test_3 |
| 1 | Importance of Data Analysis and Critical Thinking | Statistical Literacy | Teaching Approaches to Data Analysis |
| 2 | Teaching Methods and Resources | Teaching Tools | Access to Data |
| 3 | Undergraduate Instruction and Mentoring | Diversifying the Field | Software and Tools for Data Analysis |
| 4 | Graphics and Visualization | Collaboration | Teaching with Data as a Pedagogical Theme |
| 5 | Geospatial Data | Challenges in Teaching | Quantitative Research Design |
| 6 | Programming and Technical Skills | Psychology-specific Themes | Sociology and Data Analysis |
| 7 | Statistical Literacy and Research Design | Practical Skills | Geospatial Data |

**Table 5** – Themes generated for the 'teaching' dataset with T=1

We can see in Table 5 slightly more variation, compared to what we saw in Table 4. Entirely new themes appear around e.g. psychology or sociology for example. I then decided for the 'teaching' dataset to operate on a lower temperature at 0.5 (Table 6).

| | Theme names (T=0.5) |
|---|---|

| Nr. | Test_1 | Test_2 | Test_3 |
|---|---|---|---|
| 1 | Teaching Critical Thinking and Interpretation of Data | Teaching Statistical Literacy | Statistical Literacy |
| 2 | Teaching with Data Sources and Tools | Teaching with Data | Teaching Challenges |
| 3 | Mentoring and Diversifying the Field | Mentoring and Diversifying the Field | Diversity and Inclusion |
| 4 | Teaching Statistics and Research Methods | Teaching GIS | Teaching Resources |
| 5 | Teaching Geospatial Data | Teaching Research Methods | Quantitative Methods |
| 6 | Remote Teaching and Learning | Challenges in Teaching with Data | Programming |
| 7 | Sociological Research and Data Skills | Practical Data Skills in Sociology | Remote Instruction |

**Table 6** – Themes generated for the 'teaching' dataset with T=0.5

With T=0.5, there is (as expected) more consistency. Some themes clearly emerge across Table 6 and Table 2, in particular about the issue of resources, the teaching of data analysis, the teaching GIS technology among others. It may be that T at 0.5, is acceptable to review the validity of themes in Phase 4 of TA.

**Phase 5: defining and naming themes**

Braun & Clarke (2006, p. 92) state that: "It is important that by the end of this phase you can clearly define what your themes are, and what they are not. One test for this is to see whether you can describe the scope and content of each theme in a couple of sentences.". This probably is also a phase that requires the analyst's capacity to encapsulate all the previous steps as the model does not have memory of what was done. Nonetheless, I propose to perform Phase 5 to provide the model with the list of codes names and description composing each theme and one meaningful quote for each code (without theme and theme description) and asked with a prompt to provide a summary of what they mean and a name. The examples I propose here are the one in Table 3, and one additional theme, for the gaming dataset. The prompt used is presented in Figure 11, each topic of the list is the entire set of codes for each theme, including, descriptions and one quote for each.

```
prompt = f"""
Using all the topics in the list, give a summary (in 2 sentences) and a name (5 words max) for the summary
List of topics: {", ".join(topic_list)}
"""
```

**Figure 11** – Prompt used to summarise each theme again **(Prompt 4)**

| 'gaming' | Name | Description |
|---|---|---|
| Original (Phase 3) | Gaming and Education | This group includes topics related to the use of video games for education and learning, including the benefits and challenges of using games for teaching, the importance of creating education-based games, and the potential of gamification in education. |

| Rename and Summary (Phase 5) | Games for Education and Diversity | Games have the potential to teach various skills and disciplines, and can be used to bridge the gap between different target markets. However, there is a need for more games specifically designed for educational purposes and for greater diversity and representation in the gaming industry. |
|---|---|---|
| Original (Phase 3) | Ethics in Gaming | This group includes topics related to ethical issues in gaming, such as violence, diversity, monetization, piracy, and inclusion. It also includes discussions on the responsibility of developers and publishers to balance creative expression with ethical considerations. |
| Rename and Summary (Phase 5) | Ethical Issues in Gaming | Developers and publishers need to be aware of what they put in their games, especially if kids are playing them, and context is important. Games have a responsibility to recognize and address issues such as sexism and ethical concerns. |

**Table 7** – Potential renaming and two lines description, two themes as examples

We see in Table 7 that there is some slight variation in the name of the themes, and it may be possible at this stage to operate some renaming of the themes found at Phase 3. Generally, we can see that the model can provide a synthetic description of what each example theme means (remember the model has not seen the original name of the theme nor the descriptions).

## 6. Interpretation of the results

We can now compare the results seen in the previous pages with the analysis conducted on the 'gaming' and 'teaching' datasets by the original researchers. This will allow us to evaluate the extent to which the results of the LLM TA are similar or different to the actual analysis. For the comparison I propose two main criteria: 1) the theme names generated by the LLMs are very similar to the one of the original research; 2) the description of the theme produced by the LLMs conveys a very similar meaning to that proposed by the researchers, even if the theme name differs. Original themes and descriptions are compared then to the tables (1, 2, 4 and 6) seen in previous pages.

**6.1 Gaming Dataset**

The analysis of the 'gaming' dataset had been done using deductive coding (Persico et. Al. 2017a). There are two sets of themes proposed, one related to 'perspectives' (Numbered from 1 to 4 in Table 9, including 3 sub-themes) and one set related to 'pre-defined questions' used to obtain coherence in the analysis (Persico et al., 2017a). Table 9 also reports the description of each theme extrapolated (where available) from another project report, where the thematic scheme was defined (Persico et al., 2017b). The table shows if the original theme was inferred by the model in Phase 3, Phase 4, or as a code.

| Nr. | Theme in original research | Descriptions (where available from Persico et al. 2017b) | Phase 3 (Table 1) | Phase 4 (Table 4) | Inferred as code(s)[3] in Phase 2 |
|---|---|---|---|---|---|
| 1 | Educational perspective | "whether games can – and actually do – improve learning processes, in terms of participants' motivation and engagement and/or learning outcomes." | Yes (Theme Nr. 1) | | |
| 2 | Psychological Perspective | [no clear description could be identified] | **No** | **No** | **No** |
| 3 | Ethical perspective | "As games and gameful interactions of various kinds continue to permeate various spheres of society – entertainment, education, commerce, culture – attention is increasingly turning to the ethical implications of this phenomenon. […] As such, they necessarily entail an ethical dimension, both as cultural artefacts in themselves and as elements within a social communication process." | Yes (Theme Nr. 2) | | |
| 3a | Violence and Aggression | "concerns have been expressed for some time about the possible impact this may have on players, especially among the young. […] a causal link between violent video game playing and increased aggressive or violent behaviour could have tremendous personal and social consequences." | **No** | **No** | Yes |
| 3b | Monetisation | "As the platforms for playing and distributing video games have evolved and diversified over recent decades, so have the strategies adopted for monetisation games [sic]." | Yes (Theme Nr. 3) | | |
| 3c | Identity | "Those voicing ethical concerns about games on questions such as violence in game content and identity stereotyping/marginalization often associate those concerns with an idea of the prevailing games culture, particularly the presumed predominance within that culture of a specific demographic: young white heterosexual males. […]. Researchers investigating identity in video games – and video gaming culture – have focused particularly intensely on gender issues" | Yes (Theme Nr. 6) | | |
| 4 | Sociocultural/Artistic perspective | "a [panoramic] view of the cultural, social and technological impacts of the video games industry" | Yes (Theme Nr. 4) | | |
| | **Themes related to the 'overarching question' [no clear description available in the report]** | | | | |

---

[3] For Braun & Clarke some codes may become themes and vice versa in Phase 4. Although this cannot be done by the LLM, their presence as code tells us that some aspects of the theme have been inferred.

| 5 | Mobile gaming and casual gaming | Yes (Theme Nr. 7) | | |
| 6 | Game streaming and eSports | Yes (Theme Nr. 8) | | |
| 7 | Innovation and game development | Yes (Theme Nr. 5) | | |
| 8 | Game Marketing | **No** | **No** | Yes |
| 9 | Gamer communities | Yes (Theme Nr. 11) | | |
| 10 | Regulations | **No** | Yes (Test_1) - 'Gaming and Age Restriction' | |

**Table 9** – Qualitative comparison between the original and the LLM analysisMost themes were clearly identified by the model in Phase 3 of the TA (Table 1), with similar or almost identical names, and alike descriptions. For those that were not identified in Phase 3, one additional was identified in Phase 4, albeit with a clearly different name, i.e. "Gaming and Age Restriction' instead of 'Regulations' (Nr. 10 in Table 9.) Two themes were not inferred in either Phase 3 or 4, but they can be found in the list of codes, which is one of the practices suggested by Braun & Clarke in Phase 4. One theme was not identified. These considerations deserve further scrutiny.

Violence (3a) appears only once as an LLM code 'Violence in Games' (Index 51), there is no mention of aggression across the codes generated by the model. It may be that the model has overlooked this aspect, or that the prominence given to it by the researchers was associated with their interpretation.

Marketing (8) appears in 3 LLM codes: 'Marketing of Videogames' (Index 39), 'Online Marketing' (Index 59) and Marketing and Intent (Index 80). As there are 3 codes related to marketing, it is clear the model did not infer this as a possible theme.

For the 'Psychological perspective' theme not one of the codes has the word psychology (or similar) in either the codes' names or descriptions. There may be 'similar' words, such as cognition, but they do not have much prevalence. In Phase 4, with T=1, one theme hinted at 'mental health', but we cannot equate this with a psychological perspective. The model did not infer this theme.

### 6.2 Teaching Dataset

The 'teaching' dataset original analysis includes 5 main themes, and one theme had 3 sub-themes. In Table 10 the 5 main themes are reported from the original research. The 3 subthemes fall under theme 2 and include, 'conceptual understanding', 'critical evaluation' and 'working with data/tools', they are cases of learning goals for data science. The descriptions have been extrapolated from the report but note this is my interpretation of where the authors were defining the themes.

| Nr. | Theme | Descriptions (from Curty et al., 2021) | Phase 3 (Table 2) | Phase 4 (Tables 5 and 6) | Inferred as codes in Phase 2 |
| --- | --- | --- | --- | --- | --- |
| 1 | Expected student learning outcomes and ways students engage with data | "The desire to develop **critical thinking skills** and advance students' data literacy was consistently expressed at a high level across interviews. […]" (p. 6) | Yes (Theme Nr. 1, in particular the description mentions critical thinking) | | |
| 2 | Evidence of Learning Goals in Instructional Praxis | "Working with data and/or tools comprises students' abilities to engage directly with raw data sets, identifying and selecting existing data sources, gathering, managing, and manipulating data, as well as operating (at least at a basic level) tools that can help them to produce analyses and data visualizations." (p. 8) | Unclear (maybe covered by Nr 3, 5 and 6) | | |
| 3 | Main Challenges of Teaching with Data | [could not identify a clear definition, but this encompasses ethics, data use and re-use etc.] | Yes (Theme Nr. 5, the description mentioned the challenges related to teaching methods and data use) | | |
| 4 | Instructors' Training and Resource Sharing | "Instructional training and resource sharing varied among interviewees. The most common training that instructors expressed receiving on teaching with data out of their graduate education was through professional development opportunities, […]. Otherwise, the second most common method to learn the techniques to teach with data is through self-discovery" (p. 17) | Yes Nr. 7 | | |
| 5 | Types of support needed | "the need for more assistance at the university level spanning from infrastructure with better-equipped and more labs to accommodate hands-on sessions, to services and resources such as workshops on computational tools, which could be offered on an ongoing basis and considering their needs and availability." (p. 20) | Unclear (maybe subsumed in Nr. 7 by the model) | Yes (Test_3) e.g. 'Teaching Resources' | |

**Table 10** - Qualitative comparison between the original themes and the themes produced by the model

The model was perhaps less effective in inferring the themes directly by names in Phase 3 compared to the 'gaming' dataset, but the descriptions report similar ideas. Therefore, in some forms, 3 of the 5 themes emerged in Phase 3, whilst a fifth one emerged in Phase 4 (on support). The theme on 'Learning goals and praxis' did not emerge clearly. However, the model also inferred a variety of themes which do not appear in the original analysis. It may be that the focus of the analysts was on the learning process and on learning goals, however aspects such as 'mentoring' and 'collaboration' were inferred consistently by the model (Phases 3-4). Which might imply the model capacity to identify relevant themes which were not considered relevant by the analysts. In the report (Curty et al., 2021), 'mentoring' is never used as a word, and collaboration tends to refer (once) to the collaboration among staff. However, the descriptions and themes built by the model clearly relate to the students' learning.

## 7. Discussion

This paper sought to experiment on whether we can use the LLM GPT3.5-Turbo to perform an inductive Thematic Analysis of semi-structured interviews, reproducing Braun & Clarke (2006) phased approach. This paper was written as an **experiment** and as a **provocation**, largely for social sciences as an audience, but also for computer scientists working on this subject. I do not concur with the implicit uncritical views offered by other authors (e.g. Xiao et al., 2023; Gao et al., 2023) which seem to take for granted that we can do qualitative analysis with LLMs, and therefore focus on building solutions. These works follow the existing literature on using machine-learning for this kind of analysis, which however present some critical issues as highlighted for example by Baden et al. (2016), such as a strong focus on technical solutions over a methodological focus. However, approaches looking at testing whether using LLMs for qualitative analysis is viable do seem important, such as the one sketched by Schiavone et al. (2023). The work in this paper goes in this second direction. We have the evidence from this research, that it is possible to perform a qualitative analysis with LLMs, however I would recommend that these developments are tied with clear methodological discussions of what are the implications for social sciences. There is no denying that the model can infer codes and themes and even identify patterns which researchers did not consider relevant and contribute to better qualitative analysis.

The value of the **experiment** I conducted lies in the comparison with the results of the research of the 'gaming' and 'teaching' datasets. The model can infer some of the key themes of those research (which were identified comparing theme names and related descriptions). This is evident in the case of the 'gaming' datasets where the model inferred 9 of the 13 themes, at Phase 3. Most of these themes remain also with higher Temperature (Phase 4). The model never inferred as a theme the 'psychological perspective' or 'violence and aggression' which clearly had value for human analysts. For the 'teaching' dataset three themes were inferred at Phase 3 (by looking especially at the descriptions) and one at Phase 4. It is notable that for the 'teaching' dataset, the themes generated in Phase 4 present much more variety and richness. Moreover, in this case the model has inferred themes which were not considered by the original analysist, for example around students' collaboration.

This paper has been written also as a **provocation**. There already is some, albeit limited, research approaching qualitative analysis with LLMs. The provocation clearly stems from the idea of whether we can use an AI NLP model to do data analysis which is normally largely reliant on the interpretation of meaning by humans. In the end it does seem inevitable that qualitative researchers, especially in the social sciences, will have to engage with these models in ways that can help do their work better. For this, however, we would need to establish a set of methodological procedures which can ensure the quality and validity of the analysis. I offer some recommendations for this below.

### 7.1 Recommendations

This section reflects on some potential recommendations for furthering the research on using LLMs for qualitative analysis and connects these with previous literature.

**Prompting.** I agree with Xiao et al. (2023) that the generation of the right prompts is a key component for conducting qualitative analysis with LLMs. Different prompts (even aimed at the same output) often lead to different responses. The social sciences community may need to work on defining essential prompts that can produce desired analysis or establishing agreed procedures for prompting. These will need to be reported in publications as part of the methods.

**Temperature.** It would be important to agree how to use the Temperature. Whilst using T at 0 allows essentially the reproduction of the results, higher values may also contribute to validity. I suggested that it may be possible to identify the validity of themes by having the model work with e.g. T=0.5 and then verify if certain themes are repeated across different outputs. It may be possible to use statistical techniques to identify which themes are most frequent and possibly variance. These observations, however, will require further research.

**Human-AI collaboration.** I agree with Jiang et al. (2021) and Gao et al. (2023) that the likely scenario, is not one of the human analysts being replaced by AI analysts, but one of a Human-AI collaboration. However, we also need to establish what can be methodologically valid. We need to address how to keep the "Human-in-the-loop" about the decision made by the model and make room for the Human to intervene to address errors/hallucinations or increase validity. Previous research has suggested for example to use the model to generate the initial codes. I would think it may also be possible to have the model to be a second coder, i.e. to verify the validity of the analysis of the human analyst and suggest overlooked patterns.

**Phase 1 and 6.** In this experiment I argued that only the phases 2-5 of a TA are reasonably approachable with the model. For Phase 1 Gauthier & Wallace (2022) seem to suggest that actions like e.g. data cleaning amount to familiarising with the data. This is a good observation if we consider the process within the Human-AI collaboration, however just looking at the model use, this phase is not covered with the data cleaning. For Phase 6, there is debate about the use of LLMs for scientific writing (Lund et al., 2023), and its ethical implications. However, there may be a difference between having an LLM write a paper from scratch (in place of the author) and having the model write up a report about the analysis it has conducted.

**User Research.** I commend Gao et al. (2023) for having done user evaluation of the analysis with the model. Although they have done it to evaluate their tool, it does seem important that we work with the scientific community to assess the outputs of LLMs qualitative analyses. I would suggest that user research is not done just to assess software usability, but also to develop the methodological debate, around questions such as: "is this analysis valid"? "does it have the expected rigour?".

## 7.1 Issues and limitations

What has been presented here requires a recognition of limits. This was just an initial experiment, and I cannot claim it is fully comprehensive of a reproduction of an inductive TA.

**Prompting.** Building prompts producing the desired results has not been easy nor obvious. The response sometimes was inconsistent. Changing the number of themes asked to be inferred produces sometimes different themes. The results produced here are valid with the prompt used and with the proposed chunks, keeping still in mind that LLMs produce outputs based on probabilities. The interview chunks are included as csv files with this paper (or can be requested to the author), to allow for reproduction of results.

**Ethics.** I did not work with documents (e.g. book reviews), like some of the previous researchers, but with interviews. The interviews need to be analysed by a model in the cloud. Therefore, the data needs to be fully anonymised at the point of performing the analysis. It remains a grey area to understand the extent to which we can use these models on newly generated interviews. For this we would need in the future to inform respondents and obtain consent for the data to be processed by LLMs. Specific expert research will need to be done to assess all the ethical implications of using LLMs for qualitative analysis.

**Hallucination.** I found that hallucinations were produced during Phase 3, and I solved the problem by passing an index in the prompt. In one case I believe the model hallucinated with the assignation of a code to a theme. This can be seen in Table 3 where the code 'Gender and Diversity in eSports' is assigned to the Education theme. I did not correct this hallucination, as my goal is to foster discussion more than deliver solutions, but this is an important aspect which will need to be addressed at methodological level and within the Human-AI collaboration.